\documentclass[letterpaper, 10 pt, conference]{ieeeconf}  

\IEEEoverridecommandlockouts                              

\usepackage{cite}
\usepackage{amsmath,amssymb,amsfonts, nccmath}
\usepackage{amsmath}
\usepackage{hyperref}
\hypersetup{
    colorlinks=true,
    linkcolor=blue,
    citecolor= blue,
    filecolor=magenta,      
    urlcolor=blue,
}

\usepackage{cite}
\usepackage{amsmath,amssymb,amsfonts}
\usepackage{algorithmic}
\usepackage{graphicx}
\usepackage{textcomp}
\usepackage{listings}
\usepackage{xcolor}
\usepackage{caption}
\usepackage{dirtytalk} 
\def\BibTeX{{\rm B\kern-.05em{\sc i\kern-.025em b}\kern-.08em
    T\kern-.1667em\lower.7ex\hbox{E}\kern-.125emX}}

\title{\LARGE \bf
Autonomous UAV for Building Monitoring, Detection and Localisation of Faults 
}

\author{Suhas Thalanki$^{1}$, T Vijay Prashant$^{2}$, Harshith Kumar M B$^{3}$, Shayak Bhadraray$^{4}$ \\ Aravind S$^{5}$, Srikrishna BR$^{6}$, Sameer Dhole$^{7}$ 
\\ Dept of CSE, PES University$^{1,}$$^{2}$, Dept of ECE, PES University $^{3,}$$^{4,}$$^{5,}$$^{7}$, Dept of EEE, PES University $^{6}$

}

\begin{document}

\maketitle
\thispagestyle{empty}
\pagestyle{empty}

\begin{abstract}

Collapsing of structural buildings has been sighted commonly and the presence of potential faults has proved to be damaging to the buildings, resulting in accidents. It is essential to continuously monitor any building for faults where human access is restricted. With UAVs (Unmanned Aerial Vehicles) emerging in the field of computer vision, monitoring any building and detecting such faults is seen as a possibility. This paper puts forth a novel approach where an automated UAV traverses around the target building, detects any potential faults in the building, and localizes the faults. With the dimensions of the building provided, a path around the building is generated. The images captured by the onboard camera of the UAV are passed through a neural network system to confirm the presence of faults. Once a fault is detected, the UAV maneuvers itself to the corresponding position where the crack is detected. The simulation is done with ROS(Robot Operating System) using the AirSim environment which initializes ROS Wrappers and provides an integrated interface of ROS and AirSim to work with, The UAV is simulated in the same.

\par Keywords- Drones, Deep Learning, PID, Kalman Filter, Autonomous, UAV, ROS, Crack Detection, AirSim 
\end{abstract}

\section{INTRODUCTION}

\par With the rapid industrialization going across the world, the need to build strong buildings with high structural strength has become quite evident. However numerous amounts of cases of possible faults have led down to catastrophic disasters of severe structural damages to buildings \cite{c12} \cite{c13}. To prevent any possible disasters, maintenance along with continuous and accurate monitoring of buildings are necessary. Thus, the following paper involves the use of UAVs to find an efficient way of monitoring potential pitfalls in the building to avoid catastrophic hazards. 
\par The use of UAVs helps in accessing the sections of the building where human access is restricted. Further, some faults are partially visible to the naked eye and have a chance of being misinterpreted. The use of UAVs prevents such possibilities of errors. 
\par The UAV is simulated in Airsim with ROS Wrappers initialized in the AirSim simulation environment. The whole procedure is executed in two phases. The first phase is the \textit{inspection phase}, where the lift-off is to survey the whole building each layer at a time. In this process, images are captured at regular intervals. These images are fed to the neural network model which accurately detects any potential faults. Once this phase is over the UAV returns to its initial point.
\par The usage of IMU (Inertial measurement unit) is taken into consideration as the usage of GPS usually has a high chance of considerable errors near buildings \cite{c14}. To reduce the noise from sensors, a Kalman filter is implemented hence reducing errors and increasing the accuracy of co-ordinates. 
\par The second phase is the \textit{detection phase}, where the lift-off is to localize the fault which has been detected by the neural network. In addition to this, the corresponding co-ordinates for detected faults are provided. By doing so, the faults are brought to the notice of the concerned authorities, and appropriate measures are taken.

\section{LITERATURE SURVEY}

There have been various papers published which has tried to tackle this problem. \\
\par K. K. Jena, et, al. \cite{c1} proposed a system where a UAV-based bridge crack inspection model focusing on Edge Detection (ED) mechanisms is developed to support the Internet of Things (IoT) applications. The images after being taken from the UAVs are transmitted to an Information Center (IC) through the internet for processing using ED methods to identify the crack. If the crack width is greater than a threshold then actions are taken for Bridge Crack (BC) repair or Bridge Reconstruction. 
\\
\par H. Cho, et, al. \cite{c2} developed an image-based methodology for the detection of structural cracks in concrete. The system comprises an edge-based crack detection technique consisting of five steps: crack width transform, aspect ratio filtering, crack region search, hole filling, and relative thresholding.
\\
\par Further, Y. Xiao, et, al. \cite{c3} proposed a new crack detection method based on the fusion of percolation algorithm and adaptive Canny operator. The crack gray-scale image is processed by the percolation algorithm to get the binary image of crack and this binary image is named Image-A, while the crack gray-scale image is also processed by the adaptive Canny operator to obtain the binary image of crack, and this binary image is named Image-B.
\\
\par Adding to K. K. Jena's methodology, P. Giri, et, al. \cite{c4} developed a novel unmanned aerial vehicle (UAV) based inspection system to detect cracks on the lateral sides and underside of bridges. Applying obstacle avoidance modules, UAVs can reach bridges' lateral sides and underside within a safe distance. Thus, the onboard camera can capture images of the bridge surface in high resolution. To locate the positions of cracks and solve the problem of the limited horizon of a single image, a fast feature-based stitching algorithm is developed. 
\\
\par Upon a closer inspection of the above-mentioned papers, it is observed that no specific work has yet been done on localizing a crack present on the surface of a building concerning a local frame of reference. Thus we propose a solution to target this issue. The paper is organized in the format where Section III describes the methodology of the system of UAV developed, Section IV showcases Experimental results arrived at during the commencement of the system along with figures. While section V mentions the conclusions and possible future works, the paper ends with references.

\section{METHODOLOGY}

\par Given a building's plot with dimensions, a path planning algorithm is used to generate a path around the building. The building can have any dimension. The path planning algorithm, along with other algorithms like obstacle avoidance algorithm and image processing combine to form the building monitoring system. The following sections gives a brief description of the same. 
\\
\subsection{Path Planning}
The path planning algorithm plays an important role as it controls where the UAV maneuvers and also prevent random motion of the UAV. To ensure a robust method of quick maneuvering around a building, it is important to generate the shortest path for the UAV and obtain an optimal procedure of traversal along with necessary constraints such as time taken for the entire procedure and power consumption by the UAV. 
\\
\begin{figure}[htbp]
\centering
\includegraphics[width= \linewidth]{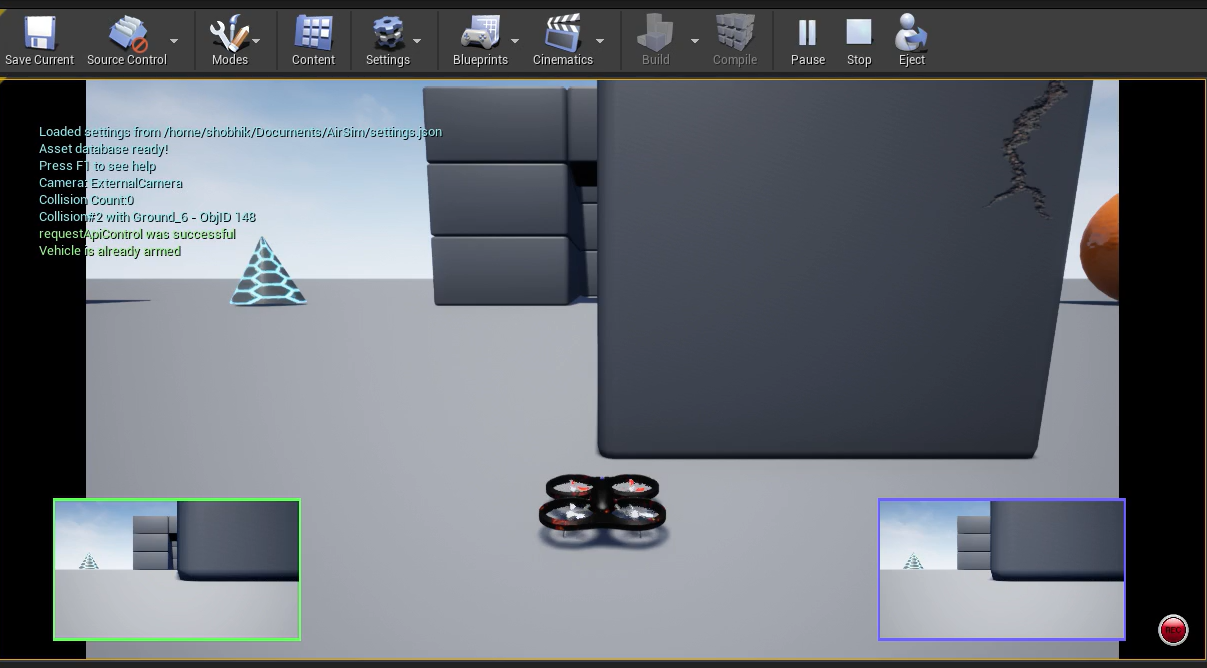}
\caption{Environment in the simulation world with UAV facing the wall always}
\label{airsim_env}
\end{figure}
\par As the UAV has to constantly face the direction towards the walls of the building when it traverses around it (As shown in Fig. \ref{airsim_env}) a search algorithm was implemented that takes the length, width and height of the building and uses the Manhattan Distance to calculate the points of traversal and generate a 2D polygon around the building. This 2D polygon is essentially the congregation of the points generated by the search algorithm, which the UAV is supposed to avoid as it covers the building itself.
\par The polygon is also given a buffer length from the perimeter of the building to act as an extra precaution for structures like the building's external accessories (balconies , veranda , railings etc.) which might act as a potential static obstacle. This buffer also ensures that the obstacle avoidance algorithm does not detect the building itself as a potential obstacle. 

   \par The UAV traverses around the building from its initial ground position to the entire path generated by the polygon. Post each iteration around the building, the UAV gains altitude and reiterates the same loop, and keeps capturing images at regular intervals until the whole building is scanned. Once the operation is complete, the UAV comes back to its home position (i.e the starting point).   
\par The neural network runs a classification test on the entire data set of pictures thus obtained to detect faults in any of the pictures and provides the co-ordinates to the faults. These co-ordinates are then taken as the destination co-ordinates and the UAV lifts off from its initial position to the defined set of co-ordinates. The shortest path from the lift-off position to the destination position is generated and the UAV maneuvers to that position through the path generated.
\subsection{PID Controller}
\par With the path being generated, the smooth maneuvering through the path generated is also essential. To stabilize the motion of the UAV, a PID  (Proportional-Integral-Differential) controller has been implemented. The PID controller is a closed feedback control system where the error is considered as the feedback. As the motion of the system is based on distance measurement, the error is assigned the value of the distance from the current position to the destination. 
\par The following are the set values of the constants used in the PID Controller system. \\ 
$K_p$ (Proportionality constant): 1.00 \\
$K_d$ (Derivative Constant): 0.5 \\ 
$K_i$ (Integral constant): 0.0001 \\ 

\par The above mentioned values were fine tuned for the the quadrotor model in AirSim and are subject to change depending on the type of UAV used.

\subsection{Obstacle Avoidance}

Obstacle avoidance is the part that is given the highest priority which perpetually runs as the background process. Once an obstacle is detected in the UAV's path, it predominates over the entire process for the control of the UAV.
\par In the process of obstacle avoidance, the closer the obstacle is detected, the greater will be the velocity buildup in the opposite direction via PID. The obstacle is detected by \textit{laserscan} rostopic in the ROS environment. It detects the obstacle and provides us with a list of distances that range from -135 degrees to 135 degrees angle.  The front of the drone is considered as 0-degree angle and a blind spot is present at the back of the drone. 

\par Hence when the obstacle distance is less than 3 meters from the obstacle, the algorithm begins to feed in the required velocities corresponding to the position of the obstacle detected, ensuring smooth maneuvering around it.

\par Once the obstacle is detected, the information received is manipulated to determine the position of the obstacle i.e., if the obstacle is on left, right or front or back or a few simple combinations of these orientations. 
\par In the case where an obstacle is detected on the right Of the UAV, the +y axis velocity is varied as the positive y-axis represents the left direction and similarly, -y-axis velocity is varied when the case of the obstacle being detected on left arises. Similarly, when the obstacle is detected on the front, it gradually moves to the left by default. But in the case where the obstacle is detected on all three sides, the x-axis direction is varied such that the drone maneuvers in the backward direction. 
\par For the varying of velocity, PID controller is implemented and inverse of the distance between the obstacle and the drone is considered as error. The mathematical expression is as follows: 
\begin{equation}
    error = \frac{1}{distance \ between \ obstacle \ and \ drone}
\end{equation}
Hence, as the UAV traverses, the distance is updated and hence the velocities are modified continuously. Once the obstacle is avoided and is no more in the peripheral region, subsequent processes take over.
\subsection{Image Processing}
\par The camera attached to the UAV along with the IMU sensors attached to it provides us with the key parameters required for the image processing algorithm to work. As soon as the path for the entire traversal of the drone is generated, it takes off, thus triggering the camera vision of the drone to be active. 
The images are captured at an interval of 10 seconds.
 \\
\par As each image is stored with a standard naming convention, the IMU Sensors keep recording the positional data of the UAV w.r.t the take-off point (acts as the origin w.r.t the local frame of reference used to generate the path). Thus, each image when captured also records the exact position of the UAV at that time and keeps the data ready in a csv file. As the traversal for the UAV ends, the images are then processed by the neural network to be classified as either a "crack" or "not a crack".
In this way, after a successful classification, we filter the co-ordinates of the drone w.r.t the images classified as "Crack" by the neural network.
\\

\par The structure made for enacting as a broken building in the present simulation uses "Asphalt Cracked" Decals as the material taken from Unreal Engine Marketplace to make a realistic simulation of a crack on a hard concrete surface. This process is shown by Fig.2\\
\begin{figure}[htbp]
\centering
\includegraphics[width= \linewidth]{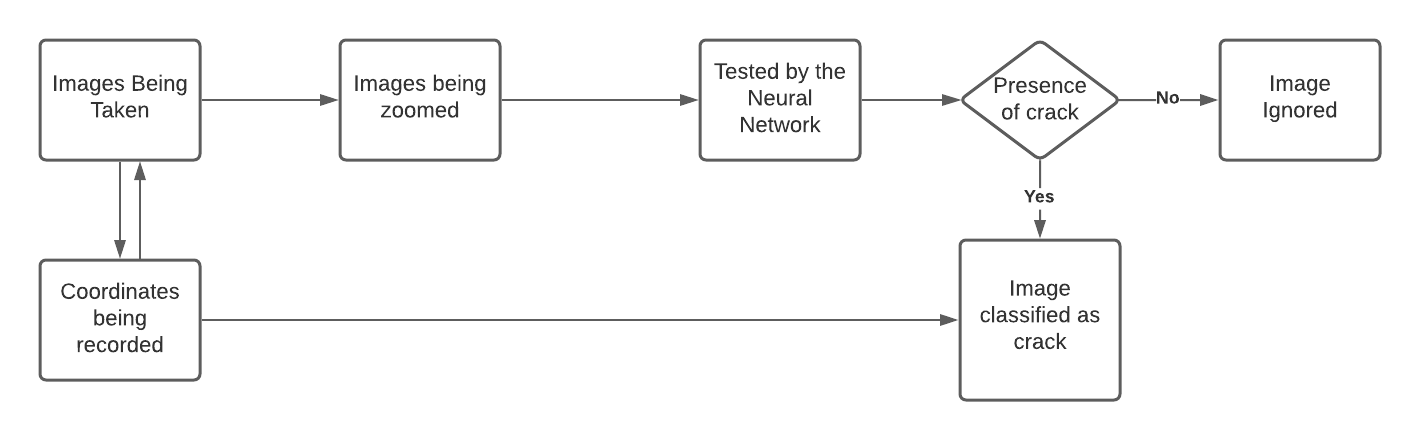}
\caption{Workflow of image processing algorithm }
\label{img_workflow}
\end{figure}
\par The Neural Network, along with the simulation resultant images was also tested on real-life images of fault lines on a building which worked out with 95 percent of accuracy.

\subsection{Complementary Filter}
\par Upon considering the use of gyroscope and accelerometer for position estimation, an important conclusion to be drawn out is that they on their own have an immense factor of noise which when used to estimate the position gives highly inaccurate values. The accelerometer is useful for long-term measurements as the noise-induced in them is minimal and the gyroscope is useful for short-term measurements as the noise-induced in them keeps getting added as time progresses. Hence the need for complementary filter design, to take the slow-moving signals from the accelerometer along with the instantaneous signals from a gyroscope and combine them to calculate the roll and pitch of the UAV arises. The gyroscope, accelerometer, and magnetometer are combined using the complementary filter design to calculate the yaw of the UAV. With the help of the Euler angles the quaternion orientation of the UAV is obtained.
\par The Gyroscope measures the angular velocity along the three major axes. So it is not directly able to predict roll, pitch, and yaw. But upon being integrated with angular velocity over time we are provided with the necessary angle, which can be used to measure the changes in roll, pitch, and yaw.

\par The third component of an IMU is the magnetometer which is capable of measuring magnetism. It helps in finding the orientation using the earth’s magnetic field, similar to how a compass works. 
\par The roll, pitch, and yaw values obtained from the accelerometer along with the magnetometer are combined with the gyroscope values to give out the necessary Euler angles. 

\subsection{Kalman Filter}
\par Kalman filtering is an algorithm that provides estimates of some unknown variables from the given measurements observed over time. Kalman filters have been proved to be useful in various applications. They have relatively simple forms and require small computational power. \\ 
Kalman filter in its most basic form consists of 3 steps.
\begin{enumerate}
    \item Predict — Based on previous knowledge about the drone position and kinematic equations, it is predicted as to what should be the position of the drone after time t+1.
    \item Measurement — Get readings from sensor regarding acceleration of the drone and compare it with Prediction 
    \item Update — Update our knowledge about the position (or state) of the drone, based on our prediction and sensor readings. 
\end{enumerate}
The values of two IMUs are used to predict the position of the drone without the use of a position determining sensor like GPS, visual odometry, etc.

\begin{figure}[htbp]
\centering
\includegraphics[width= \linewidth, height = 0.6\linewidth]{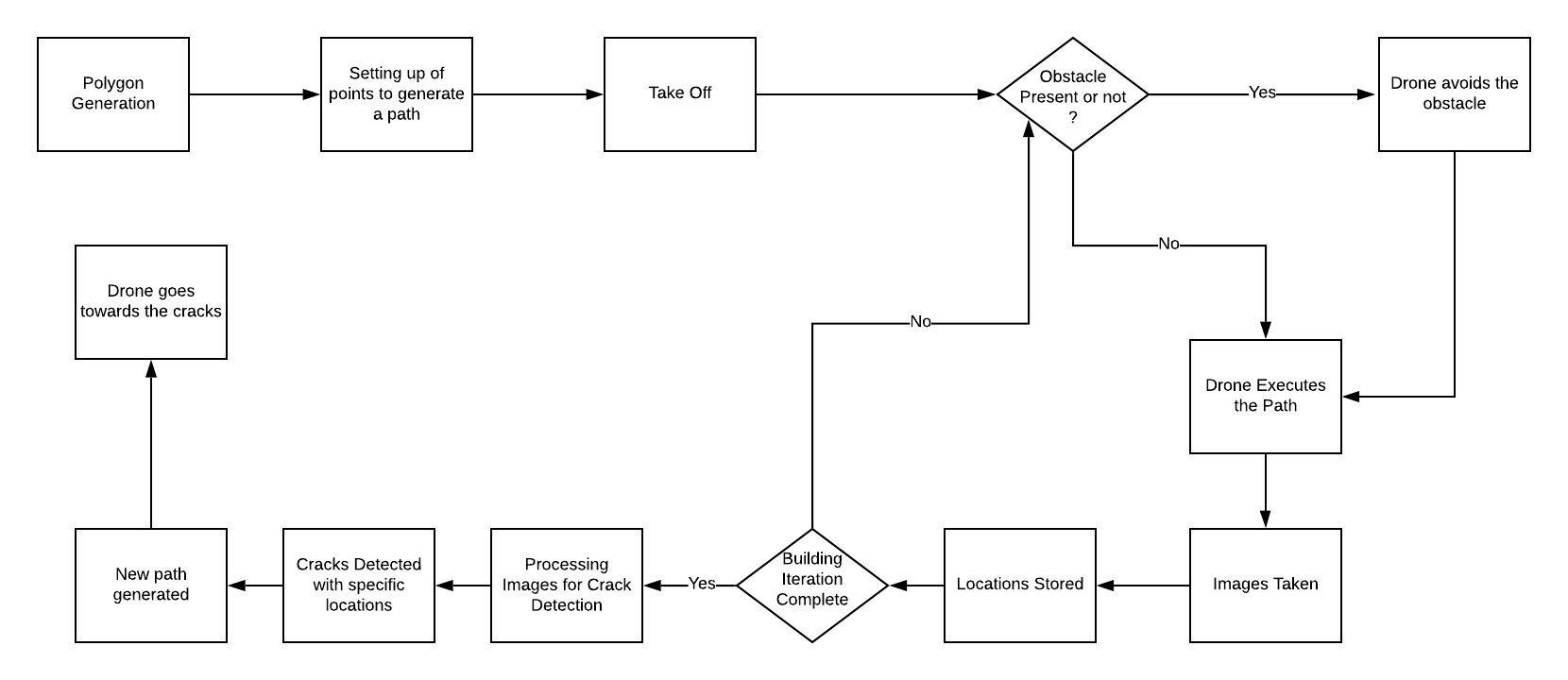}
\caption{Work flow of UAV for Fault detection}
\label{fault_detect}
\end{figure}

The complete methodology is represented by the flowchart in Fig. \ref{fault_detect}. The step-by-step procedures are essential with parallelism acting on the UAV system.

\section{Experimental Results}
The final results provide us the necessary evidence we need in support of the work done. 
 
\par
\begin{figure}[htbp]
\centering
\includegraphics[width= \linewidth]{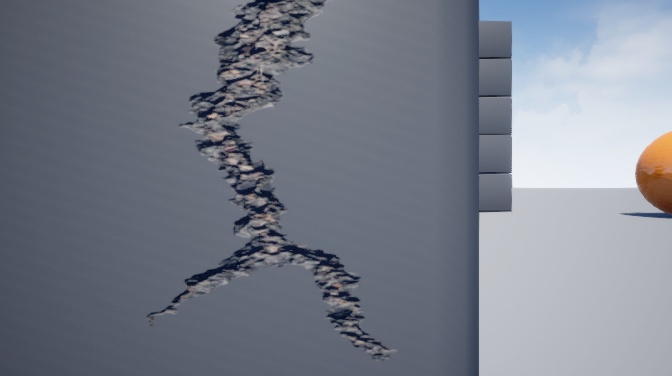}
\caption{Presence of fault line}
\label{crack_1}
\end{figure}

 The required setup for the simulation in the environment with the presence of a building with a crucial fault line on its surface was successfully achieved in AirSim. The same is shown in Fig. \ref{crack_1}

\par
\begin{figure}[htbp]
\centering
\includegraphics[width= \linewidth]{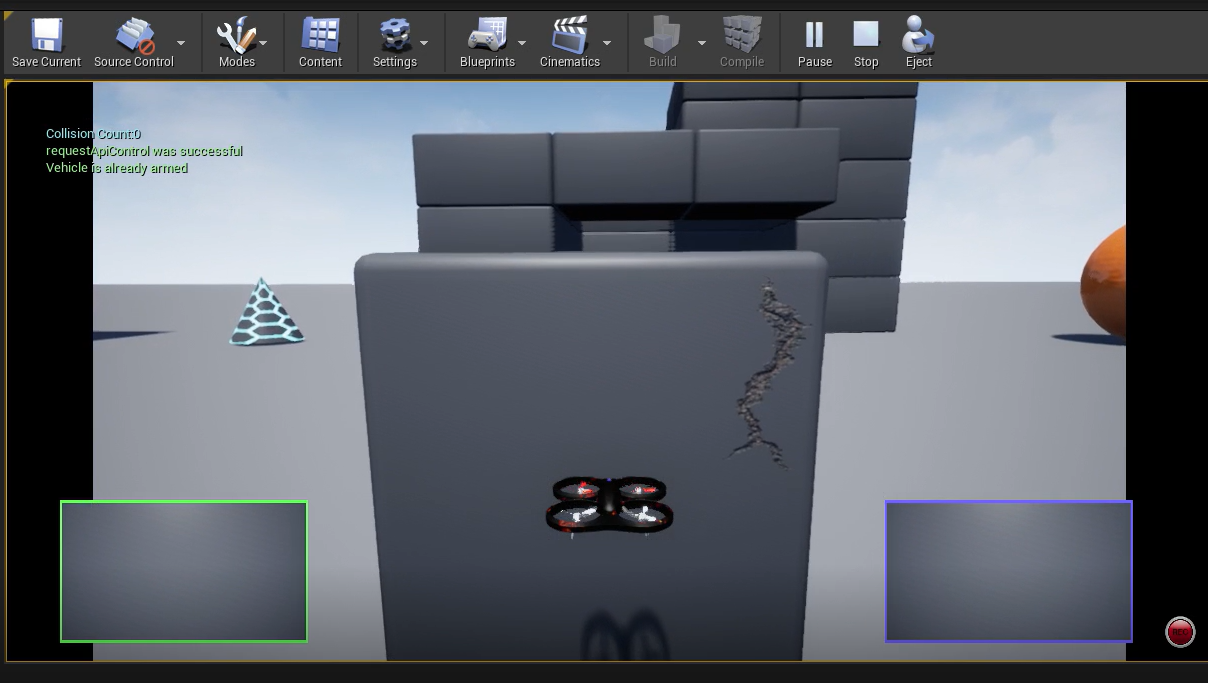}
\caption{UAV traversing the path facing the wall}
\label{trav_face_wall}
\end{figure}

\begin{figure}[htbp]
\centering
\includegraphics[width= \linewidth]{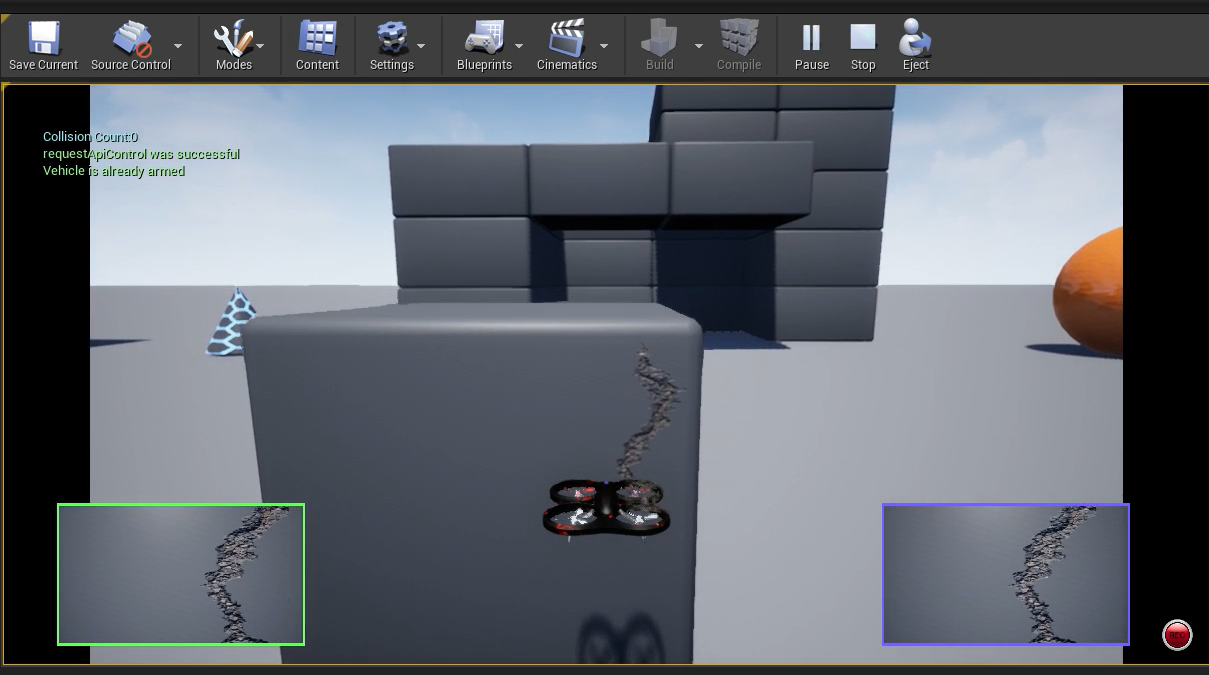}
\caption{On board camera capturing image of the fault}
\label{onboard_fault}
\end{figure}

 The UAV generates the optimum path around the perimeter of the structure, which it is to traverse around the building along with ensuring the necessary orientation to get the required images of the building at every possible angle (as shown in Fig. \ref{trav_face_wall}). The UAV, while traversing around the defined structure, also ensures a buffer region to avoid collision with any possible obstacle(static or dynamic). 
\\
While traversing the path around the building, the onboard camera keeps capturing the images along with recording the positional value of the UAV(as shown in Fig. \ref{onboard_fault}). After a successful iteration, The images are then pushed through the neural network to be classified either as cracked or not cracked. This neural network provides the results of each of the images, the corresponding co-ordinates of the fault can be filtered out from the data set obtained. 

The Fig. \ref{onboard_fault}, represents the UAV in front of the wall where the fault is seen. The onboard camera's view can also be seen on the camera window shown on left and right minimized windows. 

\par Implementation of the Kalman filter has efficiently reduced the inaccuracies in the readings of the IMU sensors to a great extent which helps in maintaining stability, manage velocities, and generating accurate positions whenever a fault is detected.  

\begin{figure}[htbp]
\centering
\includegraphics[width= \linewidth]{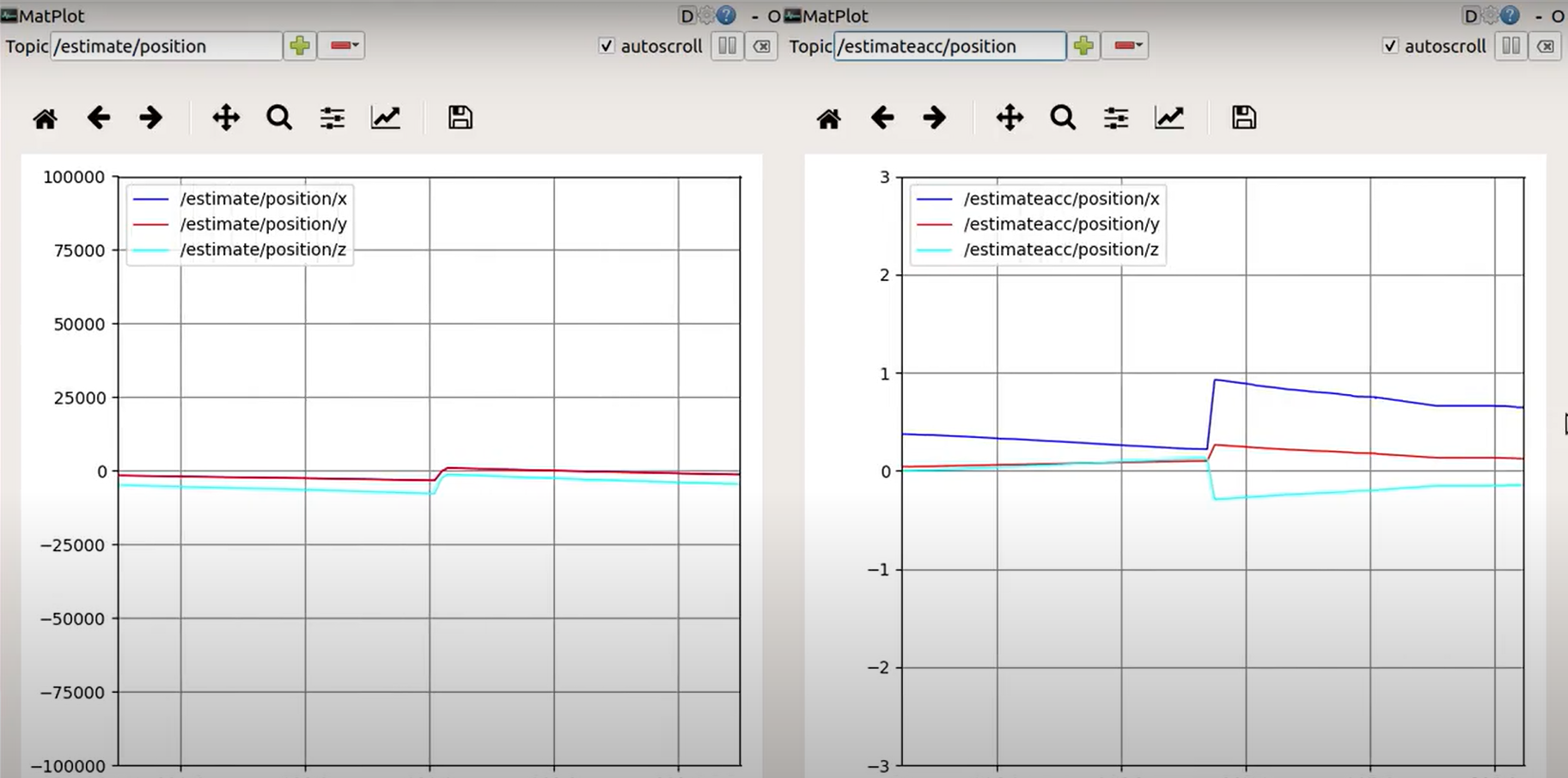}
\caption{Output of pose-estimation before and after implementation of Kalman filter}
\label{kalman}
\end{figure} 

In Fig. \ref{kalman}, the graph on left represents the pose\textunderscore estimation from the IMU data where the error oscillates from the value -25000 to +25000. Due to this, the UAV deviates from its path, and also accurate position is disturbed. Once the Kalman filter is implemented, the error oscillates from -1 to +1 which is negligible and also provides accurate IMU data and hence the position of the UAV. 

\par Thus, after obtaining the set of co-ordinates for the potential faults. The UAV traverses back to the points of interest and the same is appreciated by the camera vision of the UAV.

\par The neural network implemented for the purpose of classification and detection of cracks from the obtained image data set also provides the favourable outcome required to strengthen our claims on the working of the autonomous UAV. 
\begin{figure}[htbp]
\centering
\includegraphics[width= \linewidth]{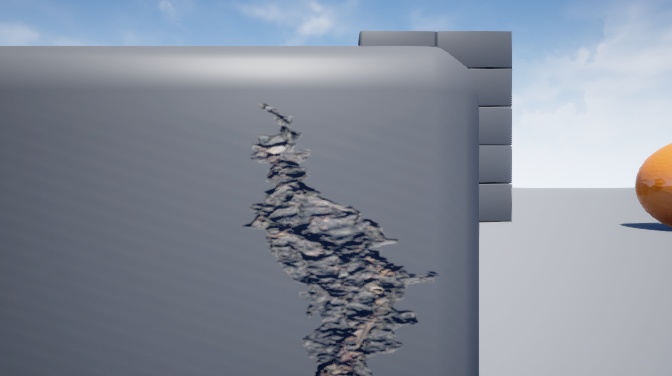}
\caption{Image classified as crack}
\label{crack_class}
\end{figure} 

\begin{figure}[htbp]
\centering
\includegraphics[width= \linewidth]{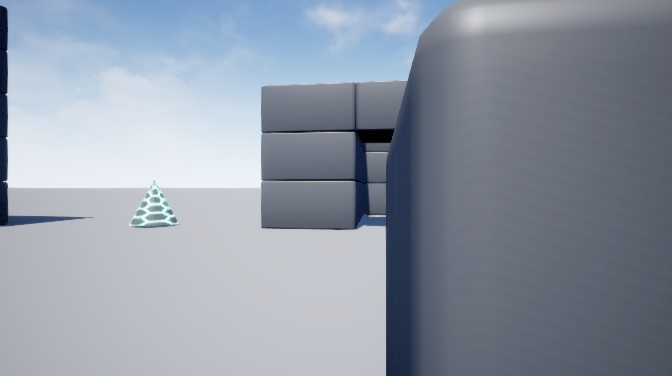}
\caption{Image classified as not cracked}
\label{nocrack_class}
\end{figure} 

\par here we can see that Fig. \ref{crack_class} successfully classified as a crack while Fig. \ref{nocrack_class} is classified as not cracked. The same neural network also subjected to a data set of similar real life images obtained by a standard camera, the classification results for the same solidifies our results in favour of our purpose.

\begin{figure}[htbp]
\centering
\includegraphics[width= \linewidth]{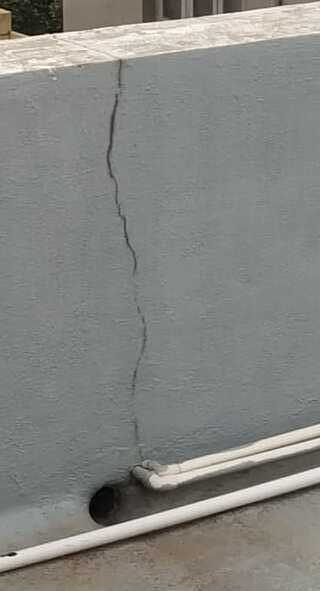}
\caption{Real life image classified as crack}
\label{real_life_crack}
\end{figure} 

\section{CONCLUSIONS AND FUTURE WORKS}
At present, the system of the automated UAV used to monitor a building and survey it for any possible faults accomplishes the same. With the UAV generating the best possible path for it to traverse around the building and also to get the best possible view of the building. It also avoids any potential obstacle present in its path by maintaining an appropriate buffer region and control over the UAV's movements. The PID controller smoothens the maneuvering across instances of both path traversal and obstacle avoidance, thus stabilizing the system. 
\par The onboard camera keeps capturing every unit of the building and hence mapping the entirety of it. Upon the successful generation of a data set with images and positional values of those instances the neural network successfully detects any potential faults. Excluding the non-significant ones. The UAV once again creates the shortest path towards the faults thus obtained, assisting to pinpoint the faults which can then be considered as hazardous and taken into immediate consideration for the required remedial procedures. Thus, avoiding a potential industrial disaster.

\par In future ventures, the present system might have the potential to be used in the crucial process of curing a specific region of interest by implementing an onboard spraying mechanism over the UAV.
\addtolength{\textheight}{-12cm}   





\end{document}